\documentclass[letterpaper, 10 pt, conference]{ieeeconf}  

\usepackage{graphicx} 
\usepackage{tabularx} 
\usepackage{svg} 
\usepackage{amsmath} 

\usepackage{graphicx}
\usepackage{amsmath}
\usepackage{amssymb}
\usepackage{booktabs}
\usepackage{wasysym}
\usepackage{textcomp}
\usepackage{stfloats}
\usepackage{url}
\usepackage{verbatim}
\usepackage{cite}
\usepackage{multirow}
\usepackage{pifont}
\usepackage{tabularx} 
\usepackage{array} 
\usepackage{multirow} 
\usepackage{xcolor} 
\usepackage[utf8]{inputenc} 
\usepackage{listings}
\usepackage{tcolorbox}

\IEEEoverridecommandlockouts                              

\title{\LARGE \bf
PromptLNet: Region-Adaptive Aesthetic Enhancement via Prompt Guidance in Low-Light Enhancement Net
}

\author{
Jun Yin\textsuperscript{1$\ast$},
Yangfan He\textsuperscript{2$\ast$},
Miao Zhang\textsuperscript{1$\dagger$},
Pengyu Zeng\textsuperscript{1},
Tianyi Wang \textsuperscript{3},
Shuai Lu\textsuperscript{1},
Xueqian Wang \textsuperscript{1}
\thanks{\textsuperscript{1}Shenzhen International Graduate School, Tsinghua University, Shenzhen 518055, China}
\thanks{\textsuperscript{2}University of Minnesota - Twin Cities, Minnesota 55455, USA}
\thanks{\textsuperscript{3}Department of Mechanical Engineering \& Materials Science, Connecticut 06511, Yale University, USA}
\thanks{\textsuperscript{$\ast$}Corresponding author: Miao Zhang, zhangmiaotju@gmail.com}
\thanks{\textsuperscript{$\dagger$}First Author and Second Author contribute equally to this work.}
}

\begin{document}

\maketitle
\thispagestyle{empty}
\pagestyle{empty}

\begin{abstract}
Learning and improving large language models through human preference feedback has become a mainstream approach, but it has rarely been applied to the field of low-light image enhancement. Existing low-light enhancement evaluations typically rely on objective metrics (such as FID, PSNR, etc.), which often result in models that perform well objectively but lack aesthetic quality. Moreover, most low-light enhancement models are primarily designed for global brightening, lacking detailed refinement. Therefore, the generated images often require additional local adjustments, leading to research gaps in practical applications. To bridge this gap, we propose the following innovations: 1) We collect human aesthetic evaluation text pairs and aesthetic scores from multiple low-light image datasets (e.g., LOL, LOL2, LOM, DCIM, MEF, etc.) to train a low-light image aesthetic evaluation model, supplemented by an optimization algorithm designed to fine-tune the diffusion model. 2) We propose a prompt-driven brightness adjustment module capable of performing fine-grained brightness and aesthetic adjustments for specific instances or regions. 3) We evaluate our method alongside existing state-of-the-art algorithms on mainstream benchmarks.
Experimental results show that our method not only outperforms traditional methods in terms of visual quality but also provides greater flexibility and controllability, paving the way for improved aesthetic quality. 
\end{abstract}

\section{INTRODUCTION}

Computer vision is viewed as a productivity engine, enhancing societal operational efficiency in critical domains such as manufacturing \cite{wang2025mdanet, zhang2025tscnet, zhang2024retinex, li2024gagent, he2025enhancing1, he2025enhancing2}, transportation\cite{li2024voltage, li2024neural}, security\cite{zhang2024adagent }, and urban governance \cite{ma2025street, zhang2023scrnet}. Among them, Low-light enhancement technology is essential in applications such as photography, traffic monitoring, and nighttime driving. By generating clear images in low-light environments, it enables users to more accurately identify potential hazards and essential information \cite{lin2017feature, zhang2017s3fd}. Current low-light enhancement techniques are generally categorized into global and regional enhancement approaches. Global enhancement methods typically adjust pixel distributions (e.g., using curve transformations and histogram equalization) to increase brightness across the image \cite{pizer1990contrast, rahman2016adaptive}. However, these methods frequently struggle to capture the intrinsic pixel distribution, often leading to issues such as color distortion and detail loss.
With advancements in deep learning, researchers have begun incorporating semantic information into low-light image enhancement \cite{yin2023cle}. Techniques like automatically generated masks and conditional diffusion models have been developed to enhance brightness in specific regions. However, these methods often rely on manually annotated masks or uniform brightness adjustments, which lack quantitative brightness control based on human preferences, thus falling short of precise, semantic-level brightness adjustments. 

In real-world applications, where lighting conditions are often complex, objective metrics alone cannot fully capture the aesthetic quality of enhanced images, which often require additional localized adjustments. Thus, despite advancements in low-light enhancement, significant gaps remain in achieving human subjective aesthetic quality and detailed control.

\begin{figure}[htbp!]
    \centering
    \includegraphics[width=1\linewidth]{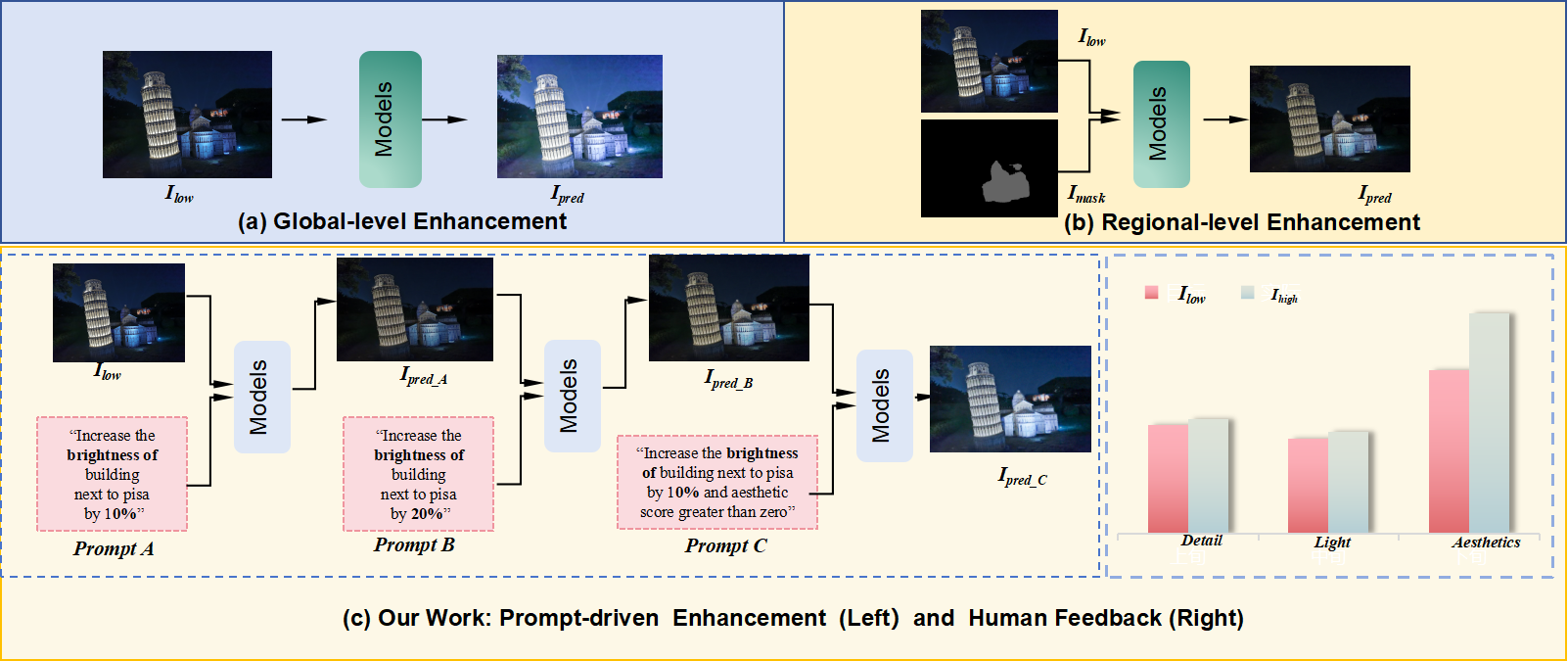}
    \caption{Overview of tasks to improve low-light conditions. The figure below illustrates the three levels of image enhancement under low-light conditions: (a) global enhancement, which is a uniform brightness adjustment of the entire image;(b) Region-level enhancement, which selectively enhances certain parts of the image, such as the background or objects of interest; (c) Prompt-driven enhancement, which uses natural language prompts to adjust the target area or the entire image under the guidance of semantic understanding. }
    \label{fig:intro}
\end{figure}

The aforementioned global light enhancement methods typically use a one-to-one mapping approach, which limits flexibility for personalized fine-tuning. In contrast, regional light enhancement algorithms enable brightness adjustments at the semantic level but still rely heavily on manually annotated masks \cite{xu2022recoro}, constraining both flexibility and scalability. Additionally, users generally prefer controlling light adjustments through natural language commands rather than manual clicking. Therefore, we propose a new prompt-driven, semantic-level approach to light enhancement. Moreover, to better align low-light enhancement with human aesthetics, we introduce reinforcement learning from human feedback (RLHF), which has demonstrated significant progress in aligning large language models with human preferences \cite{ouyang2022training}. RLHF employs a reward model (RM) to encapsulate human aesthetic standards in images, thereby guiding models to generate outputs that more closely align with human preferences. In all, we propose a prompt-driven framework for regional-level and aesthetic quality
enhancement in Low-light Environments, allowing users to  \textbf{ "Polish the Low-light Image as You Say"}.

This paper introduces the following contributions:
    \begin{enumerate}
        \item We propose a new task, \textbf{Prompt-driven Low Light Enhancement}, and a text-controlled framework that enables refined brightness and aesthetic adjustments in specific semantic instances or regions, guided by users’ natural language descriptions, thus accommodating personalized user needs.
        \item To align image enhancement with human aesthetics, we collected human aesthetic assessment text pairs from multiple low-light image datasets (e.g., LOL~\cite{wei2018deep}, LOL2, LOM, DCIM~\cite{lee2013contrast}, LIME~\cite{guo2016lime}, MEF~\cite{ma2015perceptual}) to train an aesthetic evaluation model specifically for low-light images. Based on this model, we designed an RLHF-based tuning algorithm to optimize diffusion models for better alignment with human preferences.
        \item To automate the above process, we introduce PFNet and perform a comprehensive performance evaluation. Experimental results demonstrate that our method outperforms traditional approaches in both visual quality and flexibility, opening new avenues for regional-level and aesthetic quality enhancement in low-light environments.
    \end{enumerate}

\section{The Proposed Method}

\subsection{Problem Formulation and Overall Architecture}
\label{sec:problem}
In fact, light and aesthetic adjustment requirements vary from person to person. Natural language is often the first choice to guide visual tasks at the semantic-level light enhancement and human aesthetic preferences, but its intuitiveness is difficult to translate into precise actions due to inherent ambiguity. To achieve effective control of lighting and aesthetic enhancements, two major challenges must be overcome:
\begin{itemize}
    \item \textbf{How to transform ambiguous natural language instructions into actionable quantitative conditions.}
    \item \textbf{ How to couple quantitative conditions with generative models to achieve specific effects.}
\end{itemize}

To solve these issues, we propose a novel framework using a Large Language Model (LLM) to decode abstract language and identify specific objects, their corresponding brightness levels, and aesthetic scores. Our solution unfolds in several stages: 1) Target Localization: The Retinex-based Reasoning Segment (RRS) module processes the instructions to localize the target objects (\( I_{\text{mask}} \)) for enhancement. 2) Brightness Control: The  Brightness Controllable (BC) module determines the precise lighting adjustments needed (\(M_{\text{adjustment}}\)). 3) Contextual Adaptation: The Adaptive Contextual Compensation (ACC) module incorporates additional conditions \( I_{\text{mask}} \) and \(M_{\text{adjustment}}\) into a control diffusion model, allowing for context-aware adjustments. 4)Aesthetic Adaptation Reward. This multi-stage approach leverages LLMs for interpreting text, coupled with adaptive modules for precise and context-sensitive lighting control.

\begin{figure}[htbp]
    \centering
    \includegraphics[width=1\linewidth]{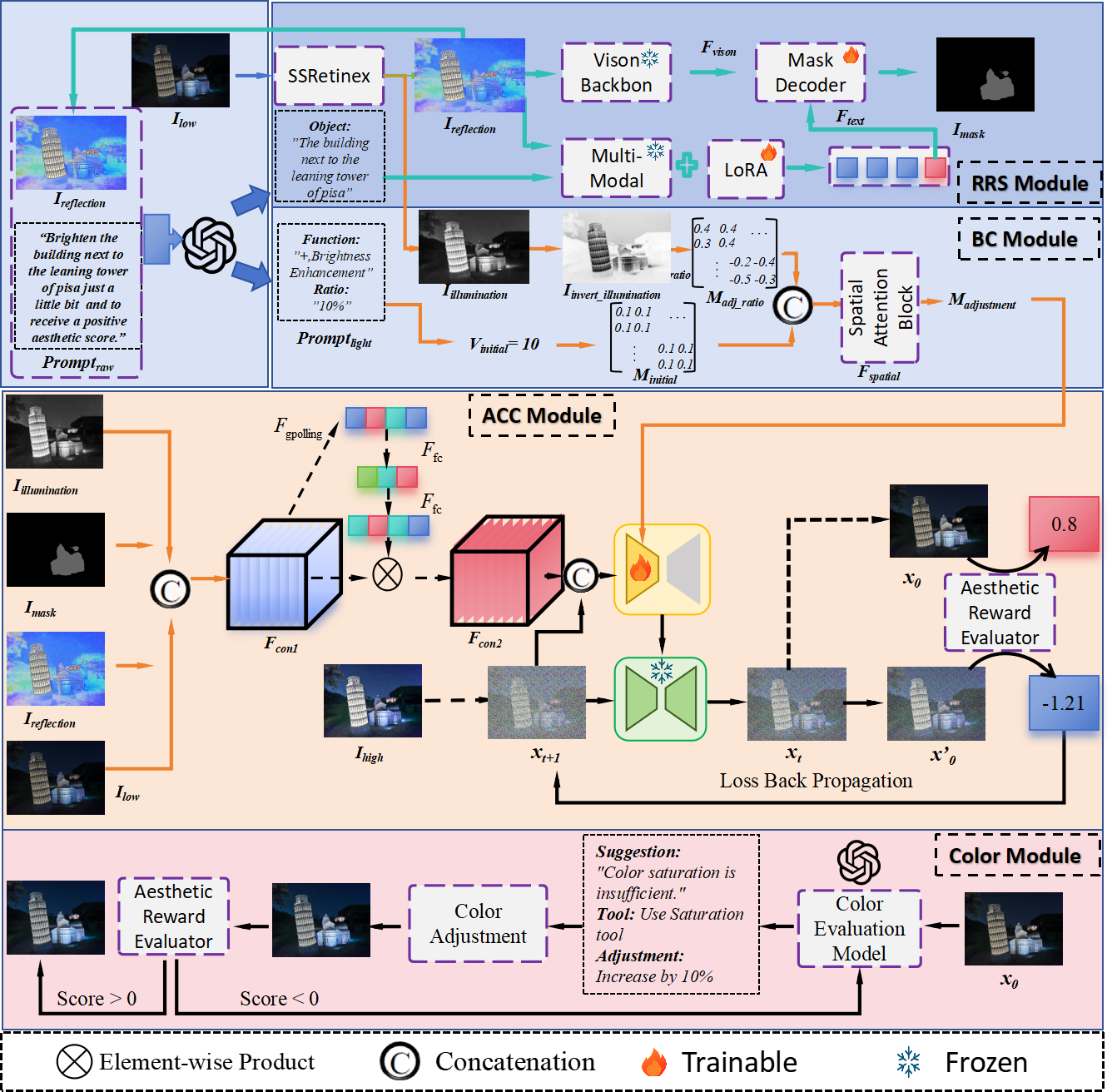}
    \caption{The overview of our framework, including the RRS Module, BC Module, ACC Module, Color Module, and control diffusion. }
    \label{fig:stage1}
    \vspace{-0.25cm}
\end{figure}

\noindent\textbf{Stage One: Target Localization and Luminance Adjustment} 

\noindent\textbf{Retinex-based Reasoning Segment (RRS) Module.} 
Accurate semantic information is crucial for semantic-level light adjustment. Traditional methods, like ReCoRo \cite{xu2022recoro} and CLE \cite{yin2023cle}, rely heavily on human annotations, which are both labor-intensive and restrict interactive flexibility. To overcome this limitation, our approach draws inspiration from \cite{zhang2024retinex} and introduces a reasoning segment powered by natural language processing to automatically identify and extract target objects from images.
The Retinex-based Reasoning Segment (RRS) module enables accurate target localization and enhancement by decomposing the input image  \( I_{\text{low}} \) into distinct components as follows:
\[
I_{\text{low}} = I_{\text{illumination}} \times I_{\text{reflection}}
\]
For more details on this fusion process, refer to \cite{yan2024visa}.
\begin{align}
 \vspace{-0.25cm}
F_{\text{fusion}} = \text{concat}(F_{\text{vision}}, F_{\text{text}})
\end{align}
The combined visual and textual features are then passed to the \textbf{Mask Decoder}\cite{lai2024lisa}, where they are used to predict the segmentation mask \( I_{\text{mask}} \):
\vspace{-0.25cm}
\begin{align}
I_{\text{mask}} = F_{\text{decoder}}(F_{\text{fusion}})
\end{align}
This approach guarantees precise pixel-level segmentation, allowing for refined low-light enhancement driven by natural language instructions.

\noindent\textbf{Brightness Control (BC) Module}
Given the variability in individual lighting preferences, precise control over light intensity is crucial. When people express their lighting needs, they often use vague language, such as "brighten it a little," without providing specific quantitative values. Most methods struggle to meet such imprecise requests. To overcome this, we leverage language models to extract quantitative parameters from these ambiguous instructions and combine them with an illumination map based on Retinex theory \cite{retinex-d4}. This enables more accurate, region-consistent lighting adjustments tailored to individual preferences.
The TBC Module controls brightness adjustments based on the natural language input \textit{\( \text{Prompt}_{\text{bright}} \)} and uses \( I_{\text{illumination}} \) as a reference for determining varying brightness levels, acting as a guide for spatial consistency. The process unfolds as follows: 1) \( I_{\text{illumination}} \) undergoes inversion, clipping, mean calculation, and normalization. 2) By utilizing a spatial attention mechanism, we generate:
\begin{align}
M_{\text{adjustment}} = 
F_{\text{spatial}}(M_{\text{initial}}, M_{\text{adj\_ratio}})
\end{align}
In summary, the TBC Module's design facilitates precise, localized brightness adjustments, with illumination maps ensuring consistent results that align closely with user specifications.

\noindent\textbf{Stage Two: Contextual Adaptation and Aesthetic Optimization}
\noindent\textbf{Adaptive  Contextual Compensation (ACC) Module:}  These modules enhance the model's ability to process diverse input data, optimize weight allocation, and generate more accurate lighting dynamics, improving image brightness.
The ACC Module processes \( I_{\text{illumination}} \), \( I_{\text{mask}} \), \( I_{\text{reflection}} \), and  \(I_{\text{low}}\) components, adaptively concatenating them into a unified feature map, \( F_{\text{con2}} \), that integrates all relevant visual information. See Figure \ref{fig:stage1} for details.
\begin{align}
F_{\text{con2}} = F_{\text{adaptive}}( F_{\text{ref\_illm}}, F_{\text{ref\_mask}},I_{\text{low}}, I_{\text{mask}})
\end{align}
This results in the final prediction of the high-quality image \( I_{\text{pred}} \).
\begin{align}
I_{\text{pred}} &= F(\boldsymbol{x} ; \Theta)+Z\left(F\left(\boldsymbol{x}+Z\left(c ; \Theta_{\mathrm{z} 1}\right) ; \Theta_{\mathrm{c}}\right) ; \Theta_{\mathrm{z} 2}\right) \\
c &= (F_{\text{con2}}, M_{\text{adjustment}})
\end{align}
\noindent\textbf{Controllable Denoising Module:} Diffusion models \cite{sohl2015deep} are a class of generative models that progressively reverse a noise-injection process, usually formulated as a Markov chain, to generate data from pure noise. 
To enhance sampling efficiency, DDIM \cite{song2020denoising} proposes a deterministic method, defined as:
\begin{equation}
\begin{aligned}
y_{t - 1} &= \sqrt{\alpha_{t - 1}}\left(\frac{y_t - \sqrt{1 - \alpha_t} \epsilon_\theta\left(y_t, t\right)}{\sqrt{\alpha_t}}\right) + \\
&\quad \sqrt{1-\alpha_{t - 1}-\sigma_t^2} \cdot \epsilon_\theta\left(y_t, t\right)+\sigma_t \epsilon_t
\end{aligned}
\end{equation}
Here, $\sigma_t^2=\eta \cdot \beta_t$, and \( \epsilon_\theta \), typically estimated using a U-Net architecture \cite{ronneberger2015u}, predicts the noise in the image at each step. The inference process begins with sampling\( y_T \sim \mathcal{N}(0, \mathbf{I}) \), which is then progressively refined to yield the final clean image \( y_0 \)
ControlNet is employed as the central model for generating images tailored to our specific requirements. 

\noindent\textbf{Aesthetic Optimization}
\noindent\textbf{Aesthetic Reward Evaluator} 
Previous aesthetic studies have primarily focused on standard lighting conditions; however, aesthetic standards vary significantly under different lighting environments. Inspired by the ImageReward approach, we construct a custom lighting-based aesthetic dataset (details provided in the attachment) and use BLIP~\cite{li2022blip} as the backbone. We extract image and text features, fuse them through cross-attention, and employ an MLP to generate a scalar value for preference comparison (details provided in the attachment). Following previous works~\cite{ouyang2022training}, we formulate the preference annotations as rankings. Specifically, we have $k \in [4,9]$ images ranked for the same prompt $T$ (with the best to worst images denoted as $x_1, x_2, ..., x_k$), yielding up to $C_k^2$ comparison pairs when there are no ties between images.
For each comparison, if $x_i$ is better and $x_j$ is worse, the loss function can be formulated as:
\begin{equation} \label{eq:loss}
\begin{split}
    \textrm{loss}(\theta) = - \mathbb{E}_{(T, x_i, x_j) \sim \mathcal{D}}[\mathop{\log}(\sigma(f_\theta(T, x_i) - f_\theta(T, x_j)))]
\end{split}
\end{equation}
where $f_\theta(T,x)$ is a scalar value of preference model for prompt $T$ and generated image $x$.

\noindent\textbf{Aesthetic Reward Feedback Learning } 

Since we cannot score the noise within the diffusion denoising process, we instead directly predict $x_t \to x_0^\prime$ at a step $t$ (which differs from the real latent $x_0$ that undergoes $x_t \to x_{t+1} \to \dots \to x_0$). $x_0^\prime$ is then evaluated by the aesthetic reward evaluator and used as the backward propagation gradient for the human preference loss (see Appendix for details).

\begin{figure}[htbp]
\centering
\includegraphics[width=0.5\textwidth,keepaspectratio]{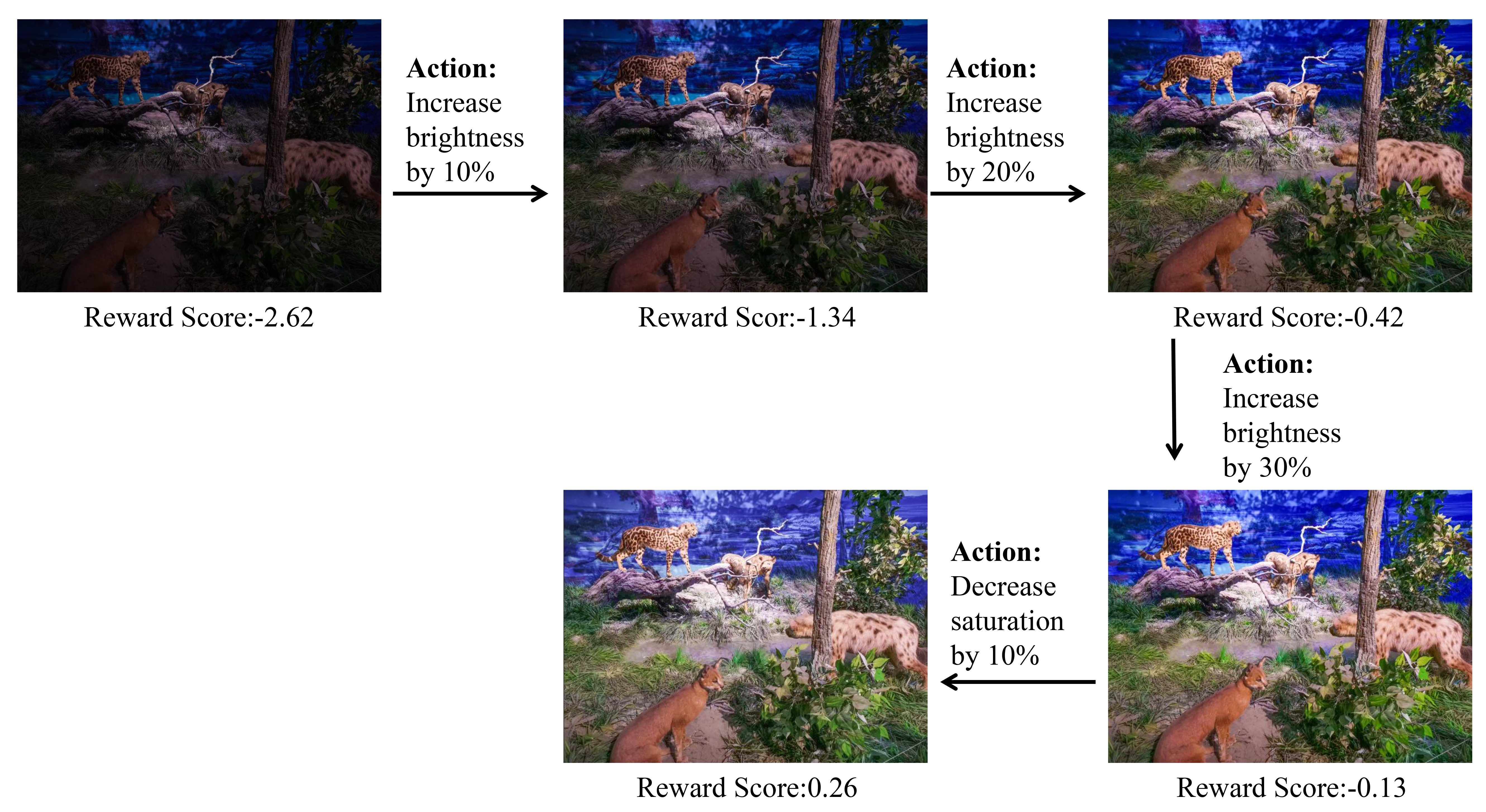}
\caption{
Step-by-step image enhancement process using iterative adjustments and reward-based evaluation in color module. The workflow begins with the original image (Reward Score: -2.82), where brightness is increased by 20\%, leading to a minor improvement in the reward score (-1.34). Subsequent adjustments include further brightening by 50\% and boosting saturation by 25\%, progressively enhancing the visual appeal while improving the reward score to -0.93. However, reducing saturation by 10\% results in a slight decline in the reward score (-1.12). Finally, a slight yellow tone adjustment achieves the optimal reward score of 0.22, indicating significant aesthetic and perceptual improvements throughout the iterative process. This demonstrates the effectiveness of the framework in balancing clarity, color quality, and aesthetic appeal. }
\vspace{-3mm}
\label{fig:finetune}
\end{figure}

\noindent\textbf{Stage Three: Color Adjustment and Aesthetic Reward}
The bottom of Figure \ref{fig:stage1} illustrates a workflow for optimizing image quality through color adjustments guided by a color evaluation model and a reward-based feedback loop. Initially, an image \( x_0 \) undergoes evaluation by a "Color Evaluation Model" with a large language model,  which assesses its color properties. Based on this assessment, the model generates a suggestion if the color quality does not meet predefined standards. For example, in this case, it recommends increasing "Color Saturation" by 10\% using a saturation adjustment tool from color adjustment toolsets. The workflow then passes the adjusted image to a "Reward Evaluator," which assigns a score based on the improved color quality. If the score is positive, indicating satisfactory color adjustment, the process is complete. However, if the score is negative, suggesting inadequate adjustment, the workflow loops back for further fine-tuning. This iterative process continues until the image meets the required quality standards, achieving a positive reward score, more details see Figure \ref{fig:finetune}.



\noindent\textbf{Loss Function}

To improve the sensitivity of our generative model, we introduce an auxiliary loss \cite{yin2023cle} that directly supervises the denoising process. This supplementary loss is formulated as:
\begin{equation}
\begin{aligned}
\mathcal{L}_{\text{aux}} = &\ \mathcal{L}_{\text{base}} + W_{\text{col}} \mathcal{L}_{\text{col}} 
& + W_{\text{ssim}} \mathcal{L}_{\text{ssim}} + W_{\text{reward}}\mathcal{L}_{\text {reward }}
\end{aligned}
\end{equation}

In this equation, \( \mathcal{L}_{\text{col}} \) represents the Angular Color Loss \cite{wang2019underexposed}, while \( \mathcal{L}_{\text{ssim}} \) is the SSIM Loss \cite{hore2010image}. The weighting factors \( W_{\text{col}}, W_{\text{reward}}, and W_{\text{ssim}}\) control the contributions of each loss term.
The Angular Color Loss is defined as:
\begin{equation}
\mathcal{L}_\text{col}=\sum_{i}{\angle\left(\hat{y_0}_i,y_i\right)},
\end{equation} 
where $i$ denotes the pixel location, and $\angle(,)$ determines the angle difference between two 3-dimensional vectors representing colors in RGB color space.

The SSIM Loss is given by:
\begin{equation}
\mathcal{L}_\text{ssim} = \frac{(2\mu_{y}\mu_{\hat{y_0}} + c_1)(2\sigma_{y\hat{y_0}} + c_2)}{(\mu_{y^2} + \mu_{\hat{y_0}}^2 + c_1)(\sigma_{y^2} + \sigma_{\hat{y_0}}^2 + c_2)},
\end{equation} 
where $\mu_y$ and $\mu_{\hat{y_0}}$ represent the mean pixel values, $\sigma_y$ and $\sigma_{\hat{y_0}}$ are the variances, and $\sigma_{y\hat{y_0}}$ is the covariance. Constants $c_1$ and $c_2$ ensure numerical stability.

The Aesthetic loss is given by:
$$
\mathcal{L}_{\text {reward}}=\mathbb{E}_{y_i \sim \mathcal{Y}}\left(\phi\left(r\left(y_i, g_\theta\left(y_i\right)\right)\right)\right)
$$
where $\theta$ denotes the parameters of the LDM, $g_\theta\left(y_i\right)$ denotes the generated image of LDM with parameters $\theta$ corresponding to prompt $y_i$. $\phi$ is reward-to-loss map function. $r$ is reward model. $\mathcal{Y}$ is prompt set.

\section{Dataset Construction}
We selected a subset of images from classic low-light enhancement datasets (including LOL, LOLv2, LOM, DCIM, LIME, NPE and MEF) and collected some natural images to form a dataset of a total of 1000 images.
\noindent\textbf{Objective}
Our goal is to create datasets under different lighting conditions and color combinations and determine the corresponding optimal image aesthetics. To achieve this goal, we developed a systematic data processing and evaluation workflow.

\noindent\textbf{Dataset Construction Steps}

\subsubsection{Light Adjustment}  

The original images are adjusted to multiple brightness levels, including a 10\%, 30\%, 100\%, and 150\% increase. This ensures that the model is capable of adapting to inputs with varying lighting conditions.  


\begin{figure}[htbp]
\centering
\includegraphics[width=0.55\textwidth,keepaspectratio]{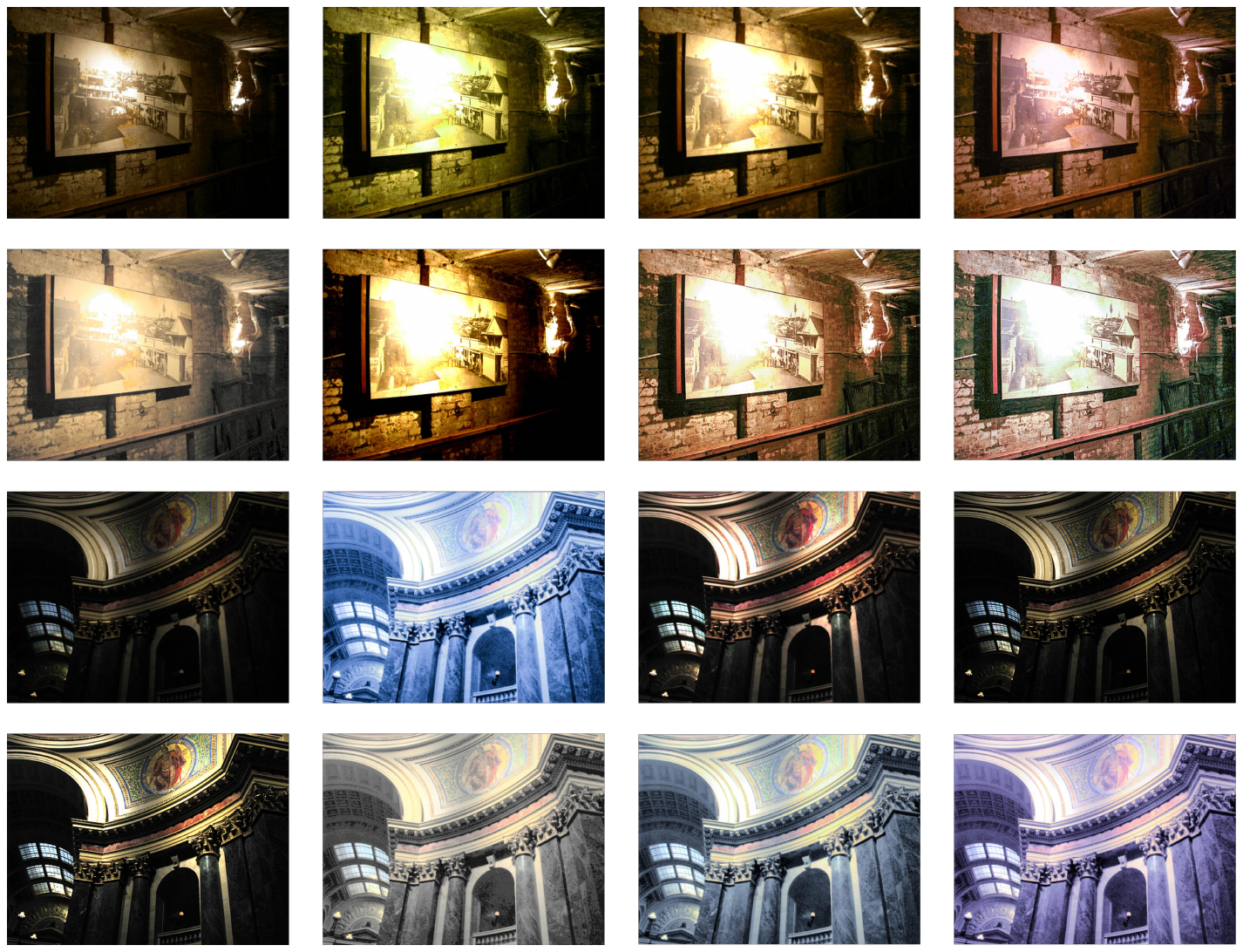}
\caption{
Examples of some images from the preprocessed DCIM and MEF datasets. These images exhibit significant differences in aspects such as hue, brightness, and saturation. }
\vspace{-3mm}
\label{fig:reward_mef}
\end{figure}

\subsubsection{Image Enhancement and Processing}

As shown in Figures \ref{fig:reward_mef}, the images undergo random operations using OpenCV tools, including the following adjustments:
\begin{itemize}
    \item \textbf{Contrast and White Balance Adjustment:} Optimization of image lighting to simulate various illumination conditions.
    \item \textbf{Sharpening and Smoothing:} Enhancing image details while controlling noise levels.
    \item \textbf{Tone and Color Grading:} Adjusting overall tone and local color to improve aesthetic performance.
    \item \textbf{Random Combination:} Randomly combining the above operations to generate a highly diverse image dataset.
\end{itemize}

\subsubsection{Scoring and Ranking}

The processed images are scored and ranked across four aesthetic dimensions. The specific workflow is as follows:
\begin{itemize}
    \item Automated preliminary scoring is performed using an \textbf{Image Reward} mechanism.
    \item Comprehensive evaluations are conducted to rank image quality across multiple dimensions.
\end{itemize}

The estimated processing scale for the experiment is:
\[
1000 \ \times 4 \ \text{brightness} \times 8 \ \text{transformations} = 32,000 \ \text{images}.
\]
The introduction of automated scoring significantly improves the efficiency of the evaluation process.

\subsection{Data Annotation and Expert Management}

In the final data annotation phase, we imposed the following strict requirements on expert annotators:
\begin{itemize}
    \item \textbf{Educational Background:} Annotators were required to have at least a university-level education.
    \item \textbf{Professional Expertise:} A thorough understanding of image aesthetics, lighting, and color theory was mandatory.
\end{itemize}

These rigorous standards ensured that the annotators possessed both the theoretical knowledge and practical expertise necessary for producing a consistent and high-quality annotated dataset. This provided a robust foundation for subsequent model training and evaluation.We recruited ten students from university art schools, specializing in fields such as sculpture, oil painting, and architecture. Their expertise provided them with a heightened sensitivity to color and visual aesthetics. Over the course of approximately one month, they completed the scoring process with precision and consistency.

\begin{figure}[htbp]
\centering
\includegraphics[width=0.5\textwidth,keepaspectratio]{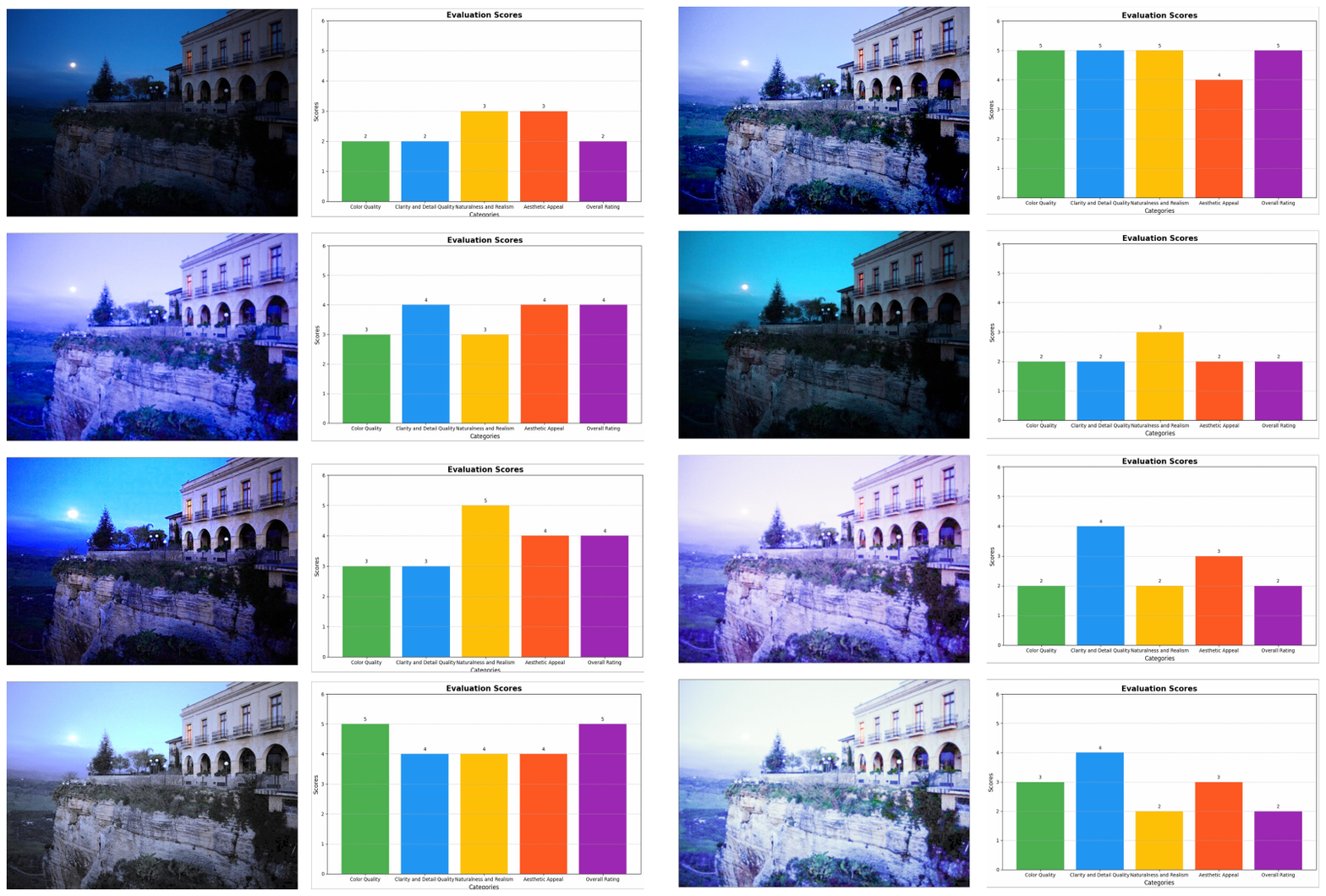}
\caption{
Examples of images from the LIME dataset along with their corresponding human feedback scores.  }
\vspace{-3mm}
\label{fig:scores}
\end{figure}

\subsection{Quantitative Measures}

\subsubsection{Color Quality}
\textbf{Definition:} The naturalness, vibrancy, and harmony of color saturation and brightness.  


\subsubsection{Clarity and Detail Quality} 

\textbf{Definition:} The sharpness of the image, the clarity of edges, and the richness of details.  

\subsubsection{Naturalness and Realism}

\textbf{Definition:} Whether the image presents a natural and realistic visual effect (avoiding over-adjustment).  


\subsection*{A.5. C. Total Score Calculation}
As the figure \ref{fig:scores} shown, scores are assigned for each dimension (1–5), and the total score is computed either by averaging or using a weighted formula.

If all five dimensions are equally weighted, the total score is calculated as:
\begin{align}
\text{Total Score} = \frac{1}{25} \big( 
& \text{Color Quality} + \text{Clarity and Detail} \notag \\
& + \text{Naturalness and Realism} \notag \\
& + \text{Aesthetic Appeal} + \text{Overall Rating} \big). \tag{15}
\end{align}

If certain dimensions (e.g., Color Quality and Clarity and Detail Quality) are more critical, the weighted score can be expressed as:
\begin{align}
\text{Total Score} = 
& w_1 \cdot \text{Color Quality} + w_2 \cdot \text{Clarity and Detail} \notag \\
& + w_3 \cdot \text{Naturalness and Realism} \notag \\
& + w_4 \cdot \text{Aesthetic Appeal} \notag \\
& + w_5 \cdot \text{Overall Rating}. \tag{16}
\end{align}

Here, \(w_1, w_2, \ldots, w_5\) represent the weights assigned to each dimension.




\section{Experimental Results}

\subsection{Experimental Setup}
The experiment proceeded in three stages: In the first RRS module stage,  we fine-tuned a pre-trained U-Net using the AdamW optimizer, with a learning rate of 0.0003 no weight decay, and a batch size of 6.
In the second aesthetic reward evaluator, fixing 70\% of transformer layers from BLIP (ViT-L for image encoder, 12-layer transformer for text encoder) with a learning rate of 1e-5 and batch size of 64.
In the third controlnet stage, the Adam optimizer was applied (initial learning rate of \(1 \times 10^{-4}\), decaying to \(1 \times 10^{-6}\) via cosine annealing), a batch size of 8, and weight decay of \(1 \times 10^{-5}\), over 1000 epochs.

\begin{figure}[htbp]
    \centering
    \includegraphics[width=1\linewidth]{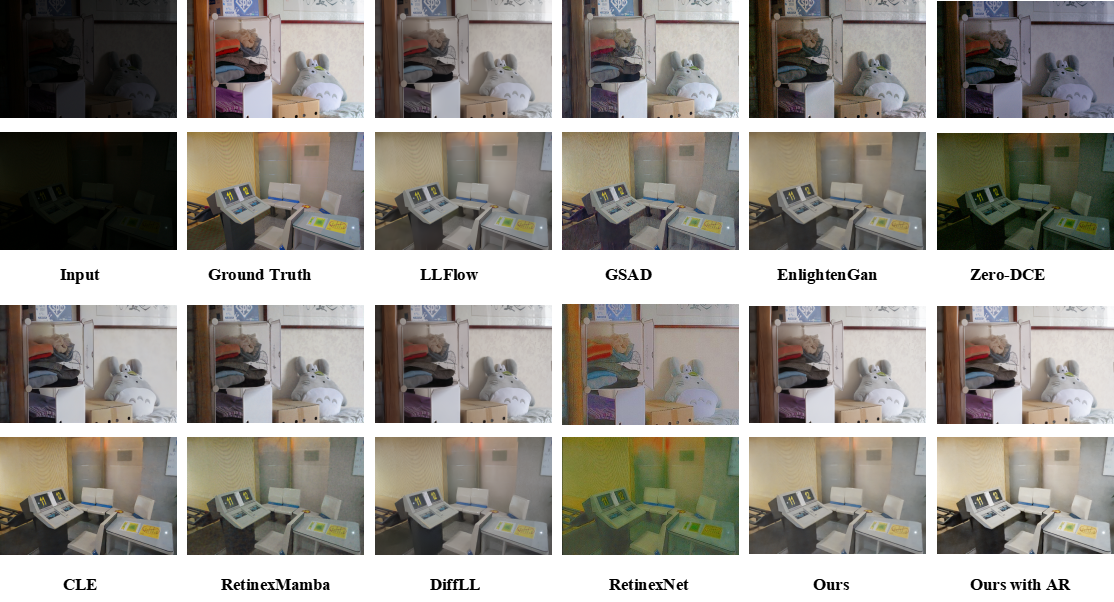}
    \caption{Visual comparison with other advanced approaches on LOL. Ours with AR means our framework with aesthetic reward (AR).}
    \label{fig:comparison_lol}
    \vspace{-0.35cm}
\end{figure}

\begin{figure}[htbp]
    \centering    
    \includegraphics[width=1\linewidth]{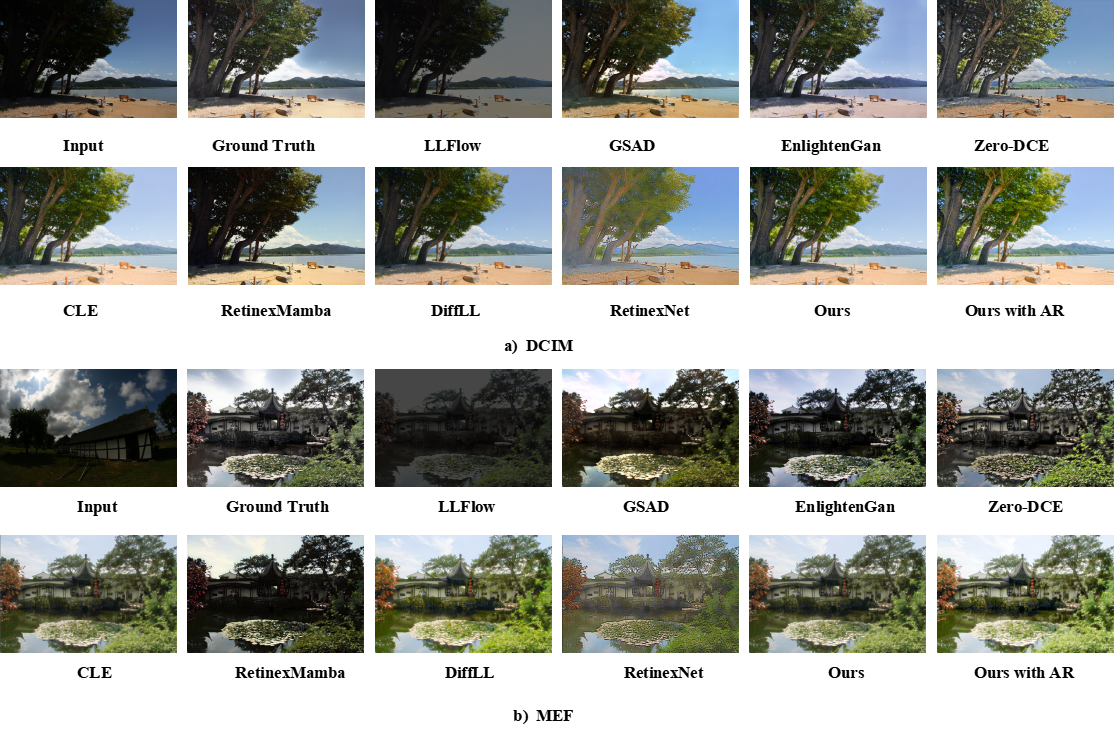}
    \caption{Visual comparison with other advanced approaches on DCIM and MEF.}
    \label{fig:comparison_lol2}
    \vspace{-0.35cm}
\end{figure}

\subsection{Datasets and Metrics}
We evaluate our model on two commonly employed benchmarks, LOL~\cite{wei2018deep} and MIT-Adobe FiveK~\cite{fivek}. The LOL dataset comprises 485 paired images for training and 15 paired images for testing, where every pair includes a low-light and a corresponding standard-light image. In the first RRS module, we generated 420 masked data pairs for training based on the LoL dataset. In the second ControlNet, to perform a quantitative task, we first use the Retinex algorithm \cite{kind_plus} to decompose the low-light image into an illumination (\( Low_{\text{illumination}} \)) image and a reflection (\( Low_{\text{reflection}} \)) image. The same process is applied to the high-light image, yielding \( High_{\text{illumination}} \) and  \( High_{\text{reflection}} \). Next, we adjust the illumination \( Low_{\text{illumination}} \) across ten levels and then combine it with the \( High_{\text{reflection}} \) to generate the ground truth image.
The MIT-Adobe FiveK dataset contains 5000 images edited by five experts using Adobe Lightroom. Following previous works~\cite{tu2022maxim,ni2020towards}, we use 4500 paired images for training and reserve 500 images for testing. To assess the quality of the output images, we use SSIM,LI-LPIPS, LPIPS~\cite{zhang2018unreasonable}, and  PSNR metrics.

\subsection{Comparison Experiment}
\begin{figure}[htbp]
    \centering    
    \includegraphics[width=1\linewidth]{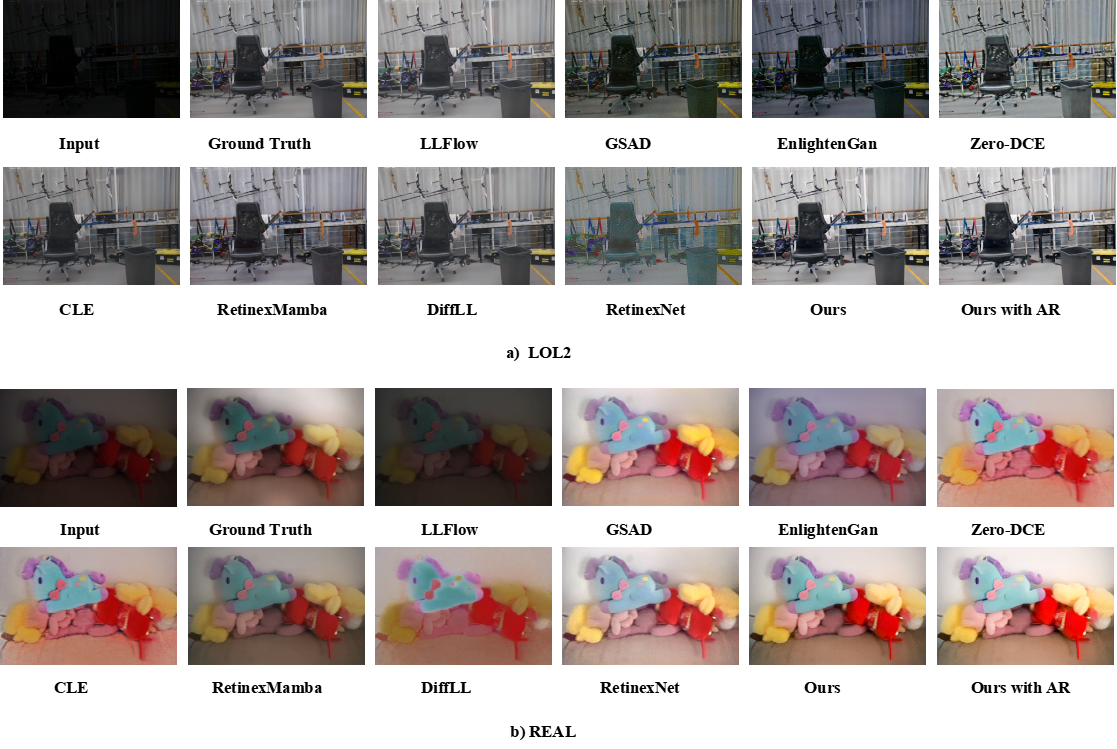}
    \caption{Visual comparison with other advanced approaches on LOL2 and REAL.}
    \label{fig:comparison_DICM}
    \vspace{-0.35cm}
\end{figure}

\noindent\textbf{Qualitative Comparison.}
Figures from \ref{fig:comparison_lol} to \ref{fig:comparison_DICM} demonstrate the effectiveness of various methods, including LLFlow~\cite{wang2022low},  GSAD~\cite{hou2024global}, EnlightenGAN~\cite{jiang2021enlightengan}, Zero-DCE~\cite{Zero-DCE}, CLE~\cite{yin2023cle}, RetinexMamba~\cite{bai2024retinexmamba},
DiffLL~\cite{jiang2023low},
Retinex-Net~\cite{wei2018deep}, our method
and our method with aesthetic reward (AR). We have demonstrated the effectiveness of our method in improving visibility in low light conditions. It has been shown that our method provides marked improvements in contrast, brightness, detail preservation without creating unnatural artifacts, whereas other methods, such as GSAD~\cite{hou2024global}  and EnlightenGAN~\cite{jiang2021enlightengan}, either overexpose or under-enhance certain areas, especially complex lighting scenarios such as DICM~\cite{lee2013contrast} and LIME~\cite{guo2016lime}.


\begin{figure}
\centering
\includegraphics[width=0.5\textwidth,keepaspectratio]{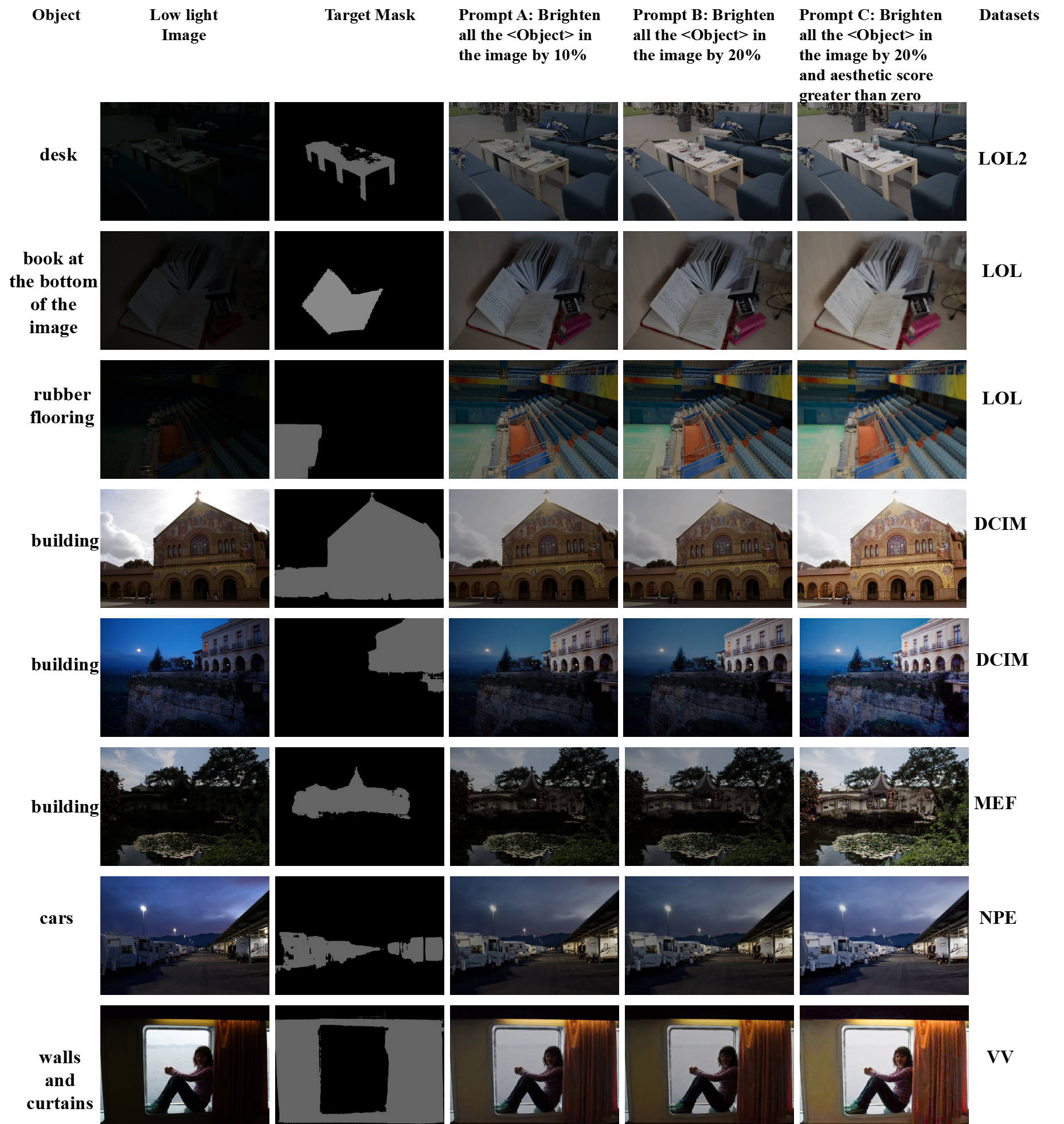}
\caption{
Targeted brightness adjustments on various objects across datasets using prompt-based control. }
\vspace{-3mm}
\label{fig:flexible_adjustment}
\end{figure}

Figure ~\ref{fig:flexible_adjustment} demonstrates our framework’s capability to precisely and selectively enhance specific regions within images based on semantic control. These examples illustrate the success of our framework in performing selective, text-driven enhancement with high precision at a semantic level.

\noindent\textbf{Quantitative Comparison.} 
Table ~\ref{tab:lol}  provides a detailed comparison of various image enhancement methods on the LOL dataset, focusing on four key metrics: Peak Signal-to-Noise Ratio (PSNR)\cite{huynh2008scope}, Structural Similarity Index Measure (SSIM)\cite{wang2004image}, Learned Perceptual Image Patch Similarity (LPIPS)\cite{zhang2018unreasonable}, and Local Illumination-LPIPS (LI-LPIPS)\cite{yin2023cle}. The performance of each method is assessed to determine its effectiveness in enhancing low-light images, with the highest result for each metric highlighted in bold and the second-best result underlined, following standard conventions in scientific reporting.
Among table, CLE achieves the highest PSNR (25.48) and LLFlow obtains the best LPIPS (0.11) and LI-LPIPS (0.1763), indicating their effectiveness in maintaining perceptual quality and reducing perceptual dissimilarities with ground truth images.
The "Ours" model, developed as part of this study, demonstrates competitive results. Furthermore, the variant "Ours with AR" shows its adaptability and robustness in enhancing image quality under different settings.
\begin{table}[t]
    \scriptsize
    \setlength\tabcolsep{4pt} 
    \renewcommand{\arraystretch}{1.2} 
    \caption{\small Comparisons on the LOL dataset. The best result is indicated in \textbf{bold}, while the second-best result is underlined.}
    \begin{tabularx}{0.48\textwidth}{>{\centering\arraybackslash}m{2.4cm}|>{\centering\arraybackslash}X>{\centering\arraybackslash}X>{\centering\arraybackslash}X>{\centering\arraybackslash}X}
        \toprule
        Method & PSNR$\uparrow$ & SSIM$\uparrow$ & LPIPS$\downarrow$ & LI-LPIPS$\downarrow$ \\
        \midrule
        Zero-DCE~\cite{guo2020zero} & 14.85 & 0.55 & 0.31 & 0.3049 \\
        EnlightenGAN~\cite{jiang2021enlightengan} & 17.49 & 0.66 & 0.31 & 0.2840 \\
        RetinexNet~\cite{Chen2018Retinex} & 16.78 & 0.57 & 0.48 & 0.5470 \\
        DRBN~\cite{yang2020fidelity} & 20.15 & 0.84 & 0.17 & 0.3269 \\
        KinD++~\cite{zhang2021beyond} & 21.29 & 0.79 & 0.16 & 0.3771 \\
        MAXIM~\cite{tu2022maxim} & 23.42 & \textbf{0.97} & 0.21 & \underline{0.1801} \\
        HWMNet~\cite{fan2022half} & 24.25 & 0.86 & 0.13 & 0.1894 \\
        LLFlow~\cite{wang2022low} & 25.19 & 0.94 & \textbf{0.11} & \textbf{0.1763} \\
        SCRnet~\cite{zhang2024retinex} & 23.17 & 0.85 & 0.21 & 0.1902 \\
        CLE~\cite{yin2023cle} & \textbf{25.48} & 0.89 & 0.16 & 0.1841 \\
        \textbf{Ours} & \underline{25.22} & \underline{0.93} & \underline{0.16} & 0.1839 \\
        \midrule
        \textbf{Ours with AR} & 24.38 & 0.91 & 0.20 & 0.1948 \\
        \bottomrule
    \end{tabularx}

\label{tab:lol}
\vspace{-3mm}
\end{table}

Table ~\ref{tab:mit} presents a comparative analysis of several image enhancement methods on the MIT-Adobe FiveK dataset."Ours" follows closely with a PSNR of 29.16 and SSIM of 0.95, demonstrating competitive performance. The "Ours with AR" variant achieves PSNR of 27.8 and SSIM of 0.89, showing slight trade-offs in quantitative metrics but maintaining robust enhancement quality.
\begin{table}[t]
    \scriptsize
    \setlength\tabcolsep{4pt} 
    \renewcommand{\arraystretch}{1.2} 
    \caption{\small Comparisons on the MIT-Adobe FiveK dataset. The best result is indicated in \textbf{bold}, while the second-best result is underlined.}
    \begin{tabularx}{0.48\textwidth}{>{\centering\arraybackslash}m{2.4cm}|>{\centering\arraybackslash}X>{\centering\arraybackslash}X}
        \toprule
        Method & PSNR$\uparrow$ & SSIM$\uparrow$ \\
        \midrule
        EnlightenGAN~\cite{jiang2021enlightengan} & 17.72 & 0.84 \\
        CycleGAN~\cite{zhu2017unpaired} & 18.24 & 0.85 \\
        Exposure~\cite{hu2018exposure} & 22.36 & 0.87 \\
        DPE~\cite{chen2018deep} & 24.06 & 0.91 \\
        UEGAN~\cite{ni2020towards} & 25.03 & 0.94 \\
        MAXIM~\cite{tu2022maxim} & 26.17 & 0.94 \\
        HWMNet~\cite{fan2022half} & 26.27 & \underline{0.95} \\
        SCRnet~\cite{zhang2024retinex} & 26.14 & 0.91 \\
        CLE~\cite{yin2023cle} & \textbf{29.81} & \textbf{0.98} \\
        \textbf{Ours} & \underline{29.16} & \underline{0.95} \\
        \midrule
        \textbf{Ours with AR} & 27.8 & 0.89 \\
        \bottomrule
    \end{tabularx}
\label{tab:mit}
\vspace{-3mm}
\end{table}

Table \ref{tab:performance_comparison} presents a comparison of various methods evaluated by aesthetic reward across seven datasets. The "Ours with AR" method shows improved performance across most datasets, achieving the highest values, which indicate alignment with human aesthetic preferences.

\begin{table}[t]
    \scriptsize
    \setlength\tabcolsep{4pt} 
    \renewcommand{\arraystretch}{1.2} 
    \caption{\small Performance comparison of aesthetic reward}
    \centering
    \begin{tabularx}{0.48\textwidth}{>{\centering\arraybackslash}m{2.4cm}|>{\centering\arraybackslash}X>{\centering\arraybackslash}X>{\centering\arraybackslash}X>{\centering\arraybackslash}X>{\centering\arraybackslash}X>{\centering\arraybackslash}X>{\centering\arraybackslash}X}
        \toprule
        \textbf{Method} & \textbf{LoL} & \textbf{LoL v2} & \textbf{LOM} & \textbf{MEF} & \textbf{DCIM} & \textbf{REAL} & \textbf{LIME} \\
        \midrule
        Original & -0.25 & -0.05 & -0.13 & -0.14 & -0.13 & -0.01 & -0.15 \\
        LLFlow & -0.23 & -0.50 & -0.47 & -0.20 & -0.10 & -0.12 & -0.02 \\
        EnlightenGan & -0.17 & -1.42 & -0.20 & -0.03 & -0.07 & -0.07 & -0.87 \\
        Zero-DCE & -0.13 & -0.09 & -0.11 & -0.19 & -0.14 & -0.07 & -0.17 \\
        KinD++ & -0.29 & -0.24 & -0.60 & -0.26 & -0.34 & -0.27 & -0.11 \\
        CLE & -0.21 & -0.35 & -0.33 & -0.27 & -1.03 & -0.08 & 0.10 \\
        RetinexMamba & 0.82 & -0.93 & 0.12 & -0.14 & 0.03 & 0.13 & 0.04 \\
        DiffLL & -0.12 & 0.02 & \textbf{0.12} & 0.18 & 0.02 & 0.07 & 0.08 \\
        RetinexNet & -0.19 & -0.01 & -0.23 & -0.08 & -0.20 & -0.23 & 0.02 \\
        \textbf{Ours} & \underline{-0.18} & -0.21 & \underline{0.07} & \underline{-0.04} & \underline{0.07} & \underline{0.14} & \underline{-0.89} \\
        \textbf{Ours with AR} & \textbf{0.20} & \textbf{0.08} & 0.02 & \textbf{0.07} & \textbf{0.16} & \textbf{0.15} & \textbf{0.26} \\
        \bottomrule
    \end{tabularx}
    \vspace{-2mm}
\label{tab:performance_comparison}
\vspace{2mm}
\end{table}


\subsection{Ablation Study}
\noindent\textbf{Ablation of Module Architecture.} 
\begin{figure}
\centering
\includegraphics[width=0.5\textwidth,keepaspectratio]{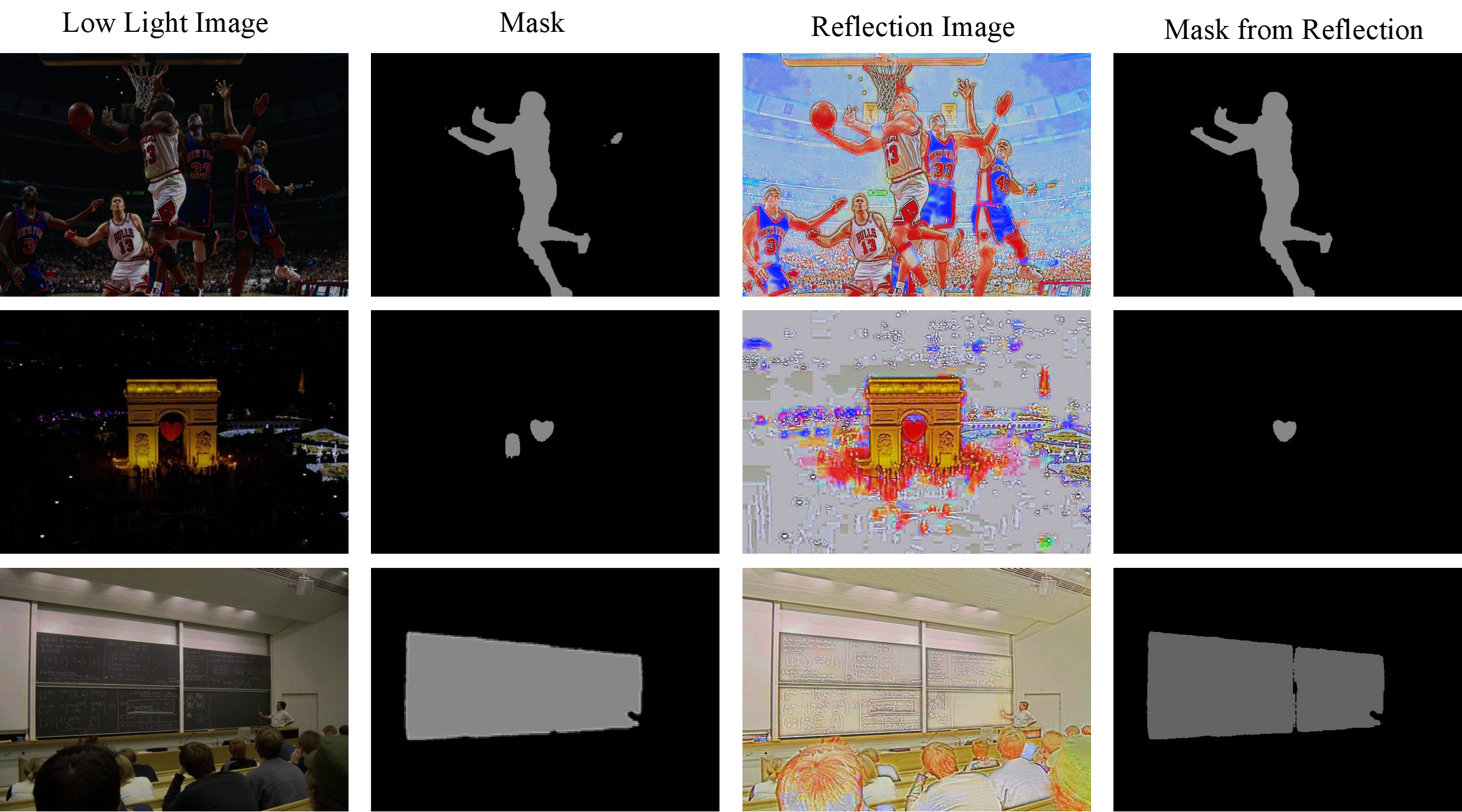}
\caption{Comparison of mask generation with low light images and reflections.}
\vspace{-8mm}
\label{fig:mask_ablation_study}
\end{figure}

\begin{table}[t]
    \scriptsize
    \setlength\tabcolsep{4pt} 
    \renewcommand{\arraystretch}{1.2} 
    \caption{\small Ablation of BC module, ACC module, and Color module  
on the LOL dataset in terms of PSNR.}
    \begin{tabularx}{0.48\textwidth}{>{\centering\arraybackslash}m{1cm}|>{\centering\arraybackslash}m{1.4cm}|>{\centering\arraybackslash}m{1.4cm}|>{\centering\arraybackslash}m{1.4cm}|>{\centering\arraybackslash}X}
        \toprule
        Case & BC Module & ACC Module & Color Module & LOL (PSNR) \\
        \midrule
        1 &  & $\checkmark$ & $\checkmark$ & 23.13 \\
        2 & $\checkmark$ &  & $\checkmark$ & 22.65 \\
        3 & $\checkmark$ & $\checkmark$ &  & 23.93 \\
        4 & $\checkmark$ & $\checkmark$ & $\checkmark$ & 25.22 \\
        \bottomrule
    \end{tabularx}
\label{tab:Ablation}
\vspace{-3mm}
\end{table}

\begin{figure}
\centering
\includegraphics[width=0.5\textwidth,keepaspectratio]{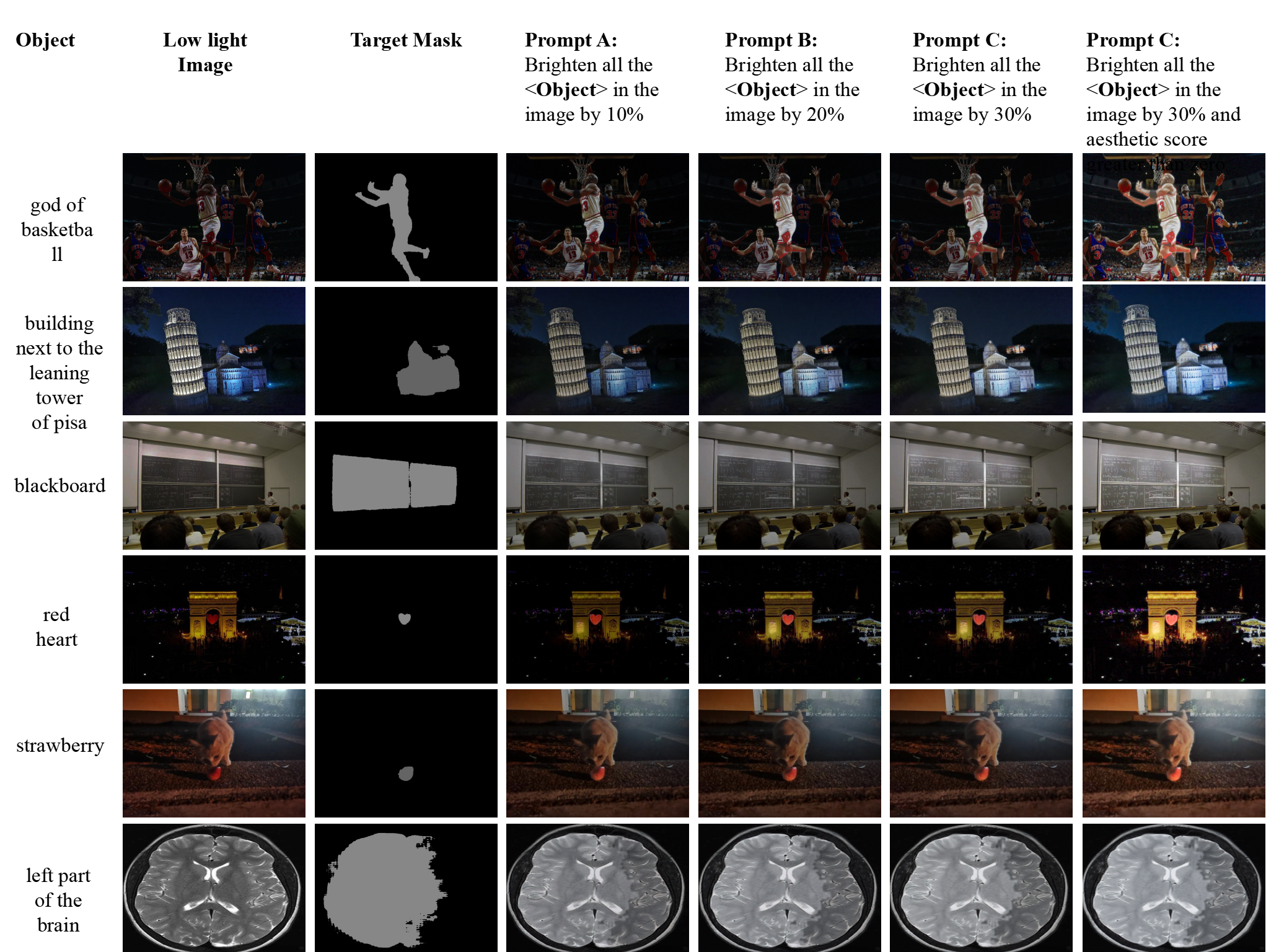}
\vspace{-5mm}
\caption{Prompt-driven lighting and aesthetic adjustments on various objects in low-light open world scenarios. }
\label{fig:local}
\end{figure}

Figure~\ref{fig:mask_ablation_study} shows the improvement in mask generation quality using RRS. To evaluate the performance of each module in the network, we conduct experiments on the LOL dataset by deleting three modules (BC, ACC and Color) respectively. Table~\ref{tab:Ablation} shows the results of ablation studies performed to measure PSNR. The PSNR of case 1 without the BC module is 23.13. The PSNR of Case 2 (including the BC module and color module) is 23.82. For Case 3 (including modules BC and ACC), the PSNR increases to 23.93. Finally, in case 4, all three modules are connected and the PSNR reaches the highest value of 25.22.

\noindent\textbf{Lighting modification and color correction in diverse, real-world scenarios.} 
Leveraging the advanced capabilities of our framework, we are able to regulate the overall image brightness to a defined level while making precise adjustments to targeted areas, all through the use of natural language. In Figure~\ref{fig:local}, these examples showcase the framework's capability to selectively modify brightness in diverse scenarios, accurately following specific text-driven instructions, even in complex open world environments.

\section{Conclusion}

In summary, this work addresses the limitations of traditional low-light image enhancement methods by incorporating human preference feedback into the evaluation process and integrating accurate text correction. Our framework improves the visual quality and aesthetics of images under low-light conditions by utilizing human aesthetic judgment data and optimizing the diffusion model. Experimental results show that this method outperforms existing methods in terms of flexibility, controllability, and practical application, and is an important step toward bridging the gap between objective measurement and aesthetic refinement in low-light image enhancement.






\bibliographystyle{IEEEtran}
\bibliography{mybib}

\begin{thebibliography}{10}
\providecommand{\url}[1]{#1}
\csname url@samestyle\endcsname
\providecommand{\newblock}{\relax}
\providecommand{\bibinfo}[2]{#2}
\providecommand{\BIBentrySTDinterwordspacing}{\spaceskip=0pt\relax}
\providecommand{\BIBentryALTinterwordstretchfactor}{4}
\providecommand{\BIBentryALTinterwordspacing}{\spaceskip=\fontdimen2\font plus
\BIBentryALTinterwordstretchfactor\fontdimen3\font minus \fontdimen4\font\relax}
\providecommand{\BIBforeignlanguage}[2]{{%
\expandafter\ifx\csname l@#1\endcsname\relax
\typeout{** WARNING: IEEEtran.bst: No hyphenation pattern has been}%
\typeout{** loaded for the language `#1'. Using the pattern for}%
\typeout{** the default language instead.}%
\else
\language=\csname l@#1\endcsname
\fi
#2}}
\providecommand{\BIBdecl}{\relax}
\BIBdecl

\bibitem{wang2025mdanet}
J.~Wang, Y.~He, K.~Li, S.~Li, L.~Zhao, J.~Yin, M.~Zhang, T.~Shi, and X.~Wang, ``Mdanet: A multi-stage domain adaptation framework for generalizable low-light image enhancement,'' \emph{Neurocomputing}, p. 129572, 2025.

\bibitem{zhang2025tscnet}
M.~Zhang, J.~Yin, P.~Zeng, Y.~Shen, S.~Lu, and X.~Wang, ``Tscnet: A text-driven semantic-level controllable framework for customized low-light image enhancement,'' \emph{Neurocomputing}, p. 129509, 2025.

\bibitem{zhang2024retinex}
M.~Zhang, Y.~Shen, Z.~Li, G.~Pan, and S.~Lu, ``A retinex structure-based low-light enhancement model guided by spatial consistency,'' in \emph{2024 IEEE International Conference on Robotics and Automation (ICRA)}.\hskip 1em plus 0.5em minus 0.4em\relax IEEE, 2024, pp. 2154--2161.

\bibitem{li2024gagent}
Z.~Li, M.~Zhang, X.~Lin, M.~Yin, S.~Lu, and X.~Wang, ``Gagent: An adaptive rigid-soft gripping agent with vision language models for complex lighting environments,'' \emph{arXiv preprint arXiv:2403.10850}, 2024.

\bibitem{he2025enhancing1}
Y.~He, J.~Wang, K.~Li, Y.~Wang, L.~Sun, J.~Yin, M.~Zhang, and X.~Wang, ``Enhancing intent understanding for ambiguous prompts through human-machine co-adaptation,'' \emph{arXiv preprint arXiv:2501.15167}, 2025.

\bibitem{he2025enhancing2}
Y.~He, S.~Li, K.~Li, J.~Wang, B.~Li, T.~Shi, J.~Yin, M.~Zhang, and X.~Wang, ``Enhancing low-cost video editing with lightweight adaptors and temporal-aware inversion,'' \emph{arXiv preprint arXiv:2501.04606}, 2025.

\bibitem{li2024voltage}
X.~Li, M.~Yang, M.~Zhang, Y.~Qi, Z.~Li, S.~Yu, Y.~Wang, L.~Shen, and X.~Li, ``Voltage regulation in polymer electrolyte fuel cell systems using gaussian process model predictive control,'' in \emph{2024 IEEE/RSJ International Conference on Intelligent Robots and Systems (IROS)}.\hskip 1em plus 0.5em minus 0.4em\relax IEEE, 2024, pp. 11\,456--11\,461.

\bibitem{li2024neural}
X.~Li, M.~Yang, Y.~Qi, and M.~Zhang, ``Neural network based model predictive control of voltage for a polymer electrolyte fuel cell system with constraints,'' \emph{arXiv preprint arXiv:2406.16871}, 2024.

\bibitem{zhang2024adagent}
M.~Zhang, Y.~Shen, J.~Yin, S.~Lu, and X.~Wang, ``Adagent: Anomaly detection agent with multimodal large models in adverse environments,'' \emph{IEEE Access}, 2024.

\bibitem{ma2025street}
X.~Ma, T.~Zeng, M.~Zhang, P.~Zeng, B.~Lin, and S.~Lu, ``Street microclimate prediction based on transformer model and street view image in high-density urban areas,'' \emph{Building and Environment}, vol. 269, p. 112490, 2025.

\bibitem{zhang2023scrnet}
M.~Zhang, Y.~Shen, and S.~Zhong, ``Scrnet: a retinex structure-based low-light enhancement model guided by spatial consistency,'' \emph{arXiv preprint arXiv:2305.08053}, 2023.

\bibitem{lin2017feature}
T.-Y. Lin, P.~Doll{\'a}r, R.~Girshick, K.~He, B.~Hariharan, and S.~Belongie, ``Feature pyramid networks for object detection,'' in \emph{Proceedings of the IEEE conference on computer vision and pattern recognition}, 2017, pp. 2117--2125.

\bibitem{zhang2017s3fd}
S.~Zhang, X.~Zhu, Z.~Lei, H.~Shi, X.~Wang, and S.~Z. Li, ``S3fd: Single shot scale-invariant face detector,'' in \emph{Proceedings of the IEEE international conference on computer vision}, 2017, pp. 192--201.

\bibitem{pizer1990contrast}
S.~M. Pizer, ``Contrast-limited adaptive histogram equalization: Speed and effectiveness stephen m. pizer, r. eugene johnston, james p. ericksen, bonnie c. yankaskas, keith e. muller medical image display research group,'' in \emph{Proceedings of the first conference on visualization in biomedical computing, Atlanta, Georgia}, vol. 337, 1990, p.~2.

\bibitem{rahman2016adaptive}
S.~Rahman, M.~M. Rahman, M.~Abdullah-Al-Wadud, G.~D. Al-Quaderi, and M.~Shoyaib, ``An adaptive gamma correction for image enhancement,'' \emph{EURASIP Journal on Image and Video Processing}, vol. 2016, pp. 1--13, 2016.

\bibitem{yin2023cle}
Y.~Yin, D.~Xu, C.~Tan, P.~Liu, Y.~Zhao, and Y.~Wei, ``Cle diffusion: Controllable light enhancement diffusion model,'' in \emph{Proceedings of the 31st ACM International Conference on Multimedia}, 2023, pp. 8145--8156.

\bibitem{xu2022recoro}
D.~Xu, H.~Poghosyan, S.~Navasardyan, Y.~Jiang, H.~Shi, and Z.~Wang, ``Recoro: Region-controllable robust light enhancement with user-specified imprecise masks,'' in \emph{Proceedings of the 30th ACM International Conference on Multimedia}, 2022, pp. 1376--1386.

\bibitem{ouyang2022training}
L.~Ouyang, J.~Wu, X.~Jiang, D.~Almeida, C.~Wainwright, P.~Mishkin, C.~Zhang, S.~Agarwal, K.~Slama, A.~Ray \emph{et~al.}, ``Training language models to follow instructions with human feedback,'' \emph{Advances in neural information processing systems}, vol.~35, pp. 27\,730--27\,744, 2022.

\bibitem{wei2018deep}
C.~Wei, W.~Wang, W.~Yang, and J.~Liu, ``Deep retinex decomposition for low-light enhancement,'' \emph{arXiv preprint arXiv:1808.04560}, 2018.

\bibitem{lee2013contrast}
C.~Lee, C.~Lee, and C.-S. Kim, ``Contrast enhancement based on layered difference representation of 2d histograms,'' \emph{IEEE transactions on image processing}, vol.~22, no.~12, pp. 5372--5384, 2013.

\bibitem{guo2016lime}
X.~Guo, Y.~Li, and H.~Ling, ``Lime: Low-light image enhancement via illumination map estimation,'' \emph{IEEE Transactions on image processing}, vol.~26, no.~2, pp. 982--993, 2016.

\bibitem{ma2015perceptual}
K.~Ma, K.~Zeng, and Z.~Wang, ``Perceptual quality assessment for multi-exposure image fusion,'' \emph{IEEE Transactions on Image Processing}, vol.~24, no.~11, pp. 3345--3356, 2015.

\bibitem{yan2024visa}
C.~Yan, H.~Wang, S.~Yan, X.~Jiang, Y.~Hu, G.~Kang, W.~Xie, and E.~Gavves, ``Visa: Reasoning video object segmentation via large language models,'' \emph{arXiv preprint arXiv:2407.11325}, 2024.

\bibitem{lai2024lisa}
X.~Lai, Z.~Tian, Y.~Chen, Y.~Li, Y.~Yuan, S.~Liu, and J.~Jia, ``Lisa: Reasoning segmentation via large language model,'' in \emph{Proceedings of the IEEE/CVF Conference on Computer Vision and Pattern Recognition}, 2024, pp. 9579--9589.

\bibitem{retinex-d4}
F.~Lv, B.~Liu, and F.~Lu, ``Fast enhancement for non-uniform illumination images using light-weight cnns,'' in \emph{Proceedings of the 28th ACM International Conference on Multimedia}, 2020, pp. 1450--1458.

\bibitem{sohl2015deep}
J.~Sohl-Dickstein, E.~A. Weiss, N.~Maheswaranathan, and S.~Ganguli, ``Deep unsupervised learning using nonequilibrium thermodynamics,'' \emph{arXiv preprint arXiv:1503.03585}, 2015.

\bibitem{song2020denoising}
J.~Song, C.~Meng, and S.~Ermon, ``Denoising diffusion implicit models,'' \emph{arXiv preprint arXiv:2010.02502}, 2020.

\bibitem{ronneberger2015u}
O.~Ronneberger, P.~Fischer, and T.~Brox, ``U-net: Convolutional networks for biomedical image segmentation,'' in \emph{International Conference on Medical image computing and computer-assisted intervention}.\hskip 1em plus 0.5em minus 0.4em\relax Springer, 2015, pp. 234--241.

\bibitem{li2022blip}
J.~Li, D.~Li, C.~Xiong, and S.~Hoi, ``Blip: Bootstrapping language-image pre-training for unified vision-language understanding and generation,'' in \emph{International Conference on Machine Learning}.\hskip 1em plus 0.5em minus 0.4em\relax PMLR, 2022, pp. 12\,888--12\,900.

\bibitem{wang2019underexposed}
R.~Wang, Q.~Zhang, C.-W. Fu, X.~Shen, W.-S. Zheng, and J.~Jia, ``Underexposed photo enhancement using deep illumination estimation,'' in \emph{Proceedings of the IEEE/CVF conference on computer vision and pattern recognition}, 2019, pp. 6849--6857.

\bibitem{hore2010image}
A.~Hore and D.~Ziou, ``Image quality metrics: Psnr vs. ssim,'' in \emph{2010 20th international conference on pattern recognition}.\hskip 1em plus 0.5em minus 0.4em\relax IEEE, 2010, pp. 2366--2369.

\bibitem{fivek}
V.~Bychkovsky, S.~Paris, E.~Chan, and F.~Durand, ``Learning photographic global tonal adjustment with a database of input / output image pairs,'' in \emph{The Twenty-Fourth IEEE Conference on Computer Vision and Pattern Recognition}, 2011.

\bibitem{kind_plus}
Y.~Zhang, X.~Guo, J.~Ma, W.~Liu, and J.~Zhang, ``Beyond brightening low-light images,'' \emph{International Journal of Computer Vision}, vol. 129, pp. 1013--1037, 2021.

\bibitem{tu2022maxim}
Z.~Tu, H.~Talebi, H.~Zhang, F.~Yang, P.~Milanfar, A.~Bovik, and Y.~Li, ``Maxim: Multi-axis mlp for image processing,'' in \emph{Proceedings of the IEEE/CVF Conference on Computer Vision and Pattern Recognition}, 2022, pp. 5769--5780.

\bibitem{ni2020towards}
Z.~Ni, W.~Yang, S.~Wang, L.~Ma, and S.~Kwong, ``Towards unsupervised deep image enhancement with generative adversarial network,'' \emph{IEEE Transactions on Image Processing}, vol.~29, pp. 9140--9151, 2020.

\bibitem{zhang2018unreasonable}
R.~Zhang, P.~Isola, A.~A. Efros, E.~Shechtman, and O.~Wang, ``The unreasonable effectiveness of deep features as a perceptual metric,'' in \emph{Proceedings of the IEEE conference on computer vision and pattern recognition}, 2018, pp. 586--595.

\bibitem{wang2022low}
Y.~Wang, R.~Wan, W.~Yang, H.~Li, L.-P. Chau, and A.~Kot, ``Low-light image enhancement with normalizing flow,'' in \emph{Proceedings of the AAAI conference on artificial intelligence}, vol.~36, no.~3, 2022, pp. 2604--2612.

\bibitem{hou2024global}
J.~Hou, Z.~Zhu, J.~Hou, H.~Liu, H.~Zeng, and H.~Yuan, ``Global structure-aware diffusion process for low-light image enhancement,'' \emph{Advances in Neural Information Processing Systems}, vol.~36, 2024.

\bibitem{jiang2021enlightengan}
Y.~Jiang, X.~Gong, D.~Liu, Y.~Cheng, C.~Fang, X.~Shen, J.~Yang, P.~Zhou, and Z.~Wang, ``Enlightengan: Deep light enhancement without paired supervision,'' \emph{IEEE transactions on image processing}, vol.~30, pp. 2340--2349, 2021.

\bibitem{Zero-DCE}
C.~G. Guo, C.~Li, J.~Guo, C.~C. Loy, J.~Hou, S.~Kwong, and R.~Cong, ``Zero-reference deep curve estimation for low-light image enhancement,'' in \emph{Proceedings of the IEEE conference on computer vision and pattern recognition (CVPR)}, June 2020, pp. 1780--1789.

\bibitem{bai2024retinexmamba}
J.~Bai, Y.~Yin, and Q.~He, ``Retinexmamba: Retinex-based mamba for low-light image enhancement,'' \emph{arXiv preprint arXiv:2405.03349}, 2024.

\bibitem{jiang2023low}
H.~Jiang, A.~Luo, H.~Fan, S.~Han, and S.~Liu, ``Low-light image enhancement with wavelet-based diffusion models,'' \emph{ACM Transactions on Graphics (TOG)}, vol.~42, no.~6, pp. 1--14, 2023.

\bibitem{huynh2008scope}
Q.~Huynh-Thu and M.~Ghanbari, ``Scope of validity of psnr in image/video quality assessment,'' \emph{Electronics letters}, vol.~44, no.~13, pp. 800--801, 2008.

\bibitem{wang2004image}
Z.~Wang, A.~C. Bovik, H.~R. Sheikh, and E.~P. Simoncelli, ``Image quality assessment: from error visibility to structural similarity,'' \emph{IEEE transactions on image processing}, vol.~13, no.~4, pp. 600--612, 2004.

\bibitem{guo2020zero}
C.~Guo, C.~Li, J.~Guo, C.~C. Loy, J.~Hou, S.~Kwong, and R.~Cong, ``Zero-reference deep curve estimation for low-light image enhancement,'' in \emph{Proceedings of the IEEE/CVF Conference on Computer Vision and Pattern Recognition}, 2020, pp. 1780--1789.

\bibitem{Chen2018Retinex}
W.~Chen, W.~Wenjing, Y.~Wenhan, and L.~Jiaying, ``Deep retinex decomposition for low-light enhancement,'' in \emph{British Machine Vision Conference}.\hskip 1em plus 0.5em minus 0.4em\relax British Machine Vision Association, 2018.

\bibitem{yang2020fidelity}
W.~Yang, S.~Wang, Y.~Fang, Y.~Wang, and J.~Liu, ``From fidelity to perceptual quality: A semi-supervised approach for low-light image enhancement,'' in \emph{Proceedings of the IEEE/CVF conference on computer vision and pattern recognition}, 2020, pp. 3063--3072.

\bibitem{zhang2021beyond}
Y.~Zhang, X.~Guo, J.~Ma, W.~Liu, and J.~Zhang, ``Beyond brightening low-light images,'' \emph{International Journal of Computer Vision}, vol. 129, pp. 1013--1037, 2021.

\bibitem{fan2022half}
C.-M. Fan, T.-J. Liu, and K.-H. Liu, ``Half wavelet attention on m-net+ for low-light image enhancement,'' in \emph{2022 IEEE International Conference on Image Processing (ICIP)}.\hskip 1em plus 0.5em minus 0.4em\relax IEEE, 2022, pp. 3878--3882.

\bibitem{zhu2017unpaired}
J.-Y. Zhu, T.~Park, P.~Isola, and A.~A. Efros, ``Unpaired image-to-image translation using cycle-consistent adversarial networks,'' in \emph{Proceedings of the IEEE international conference on computer vision}, 2017, pp. 2223--2232.

\bibitem{hu2018exposure}
Y.~Hu, H.~He, C.~Xu, B.~Wang, and S.~Lin, ``Exposure: A white-box photo post-processing framework,'' \emph{ACM Transactions on Graphics (TOG)}, vol.~37, no.~2, pp. 1--17, 2018.

\bibitem{chen2018deep}
Y.-S. Chen, Y.-C. Wang, M.-H. Kao, and Y.-Y. Chuang, ``Deep photo enhancer: Unpaired learning for image enhancement from photographs with gans,'' in \emph{Proceedings of the IEEE Conference on Computer Vision and Pattern Recognition}, 2018, pp. 6306--6314.

\end{thebibliography}


\end{document}